\documentclass[10pt,twocolumn,letterpaper]{article}

\usepackage{iccv}
\usepackage{times}
\usepackage{epsfig}
\usepackage{graphicx}
\usepackage{amsmath}
\usepackage{amssymb}
\usepackage{float,wrapfig}
\usepackage{textcomp,booktabs}
\usepackage[usenames,dvipsnames]{color}
\usepackage{colortbl}
\usepackage{color}
\usepackage{subfigure}

\usepackage[breaklinks=true,bookmarks=false]{hyperref}

\iccvfinalcopy 

\def\iccvPaperID{****} 

\ificcvfinal\fi

\begin{document}


\title{Label-PEnet: Sequential Label Propagation and Enhancement Networks \\ for Weakly Supervised Instance Segmentation }

 \author{Weifeng Ge$^{1,2,3}$, Sheng Guo$^{1,2}$, Weilin Huang$^{1,2}$\thanks{Weilin Huang is the corresponding author.}, and Matthew R. Scott$^{1,2}$\\
$^{1}$Malong Technologies, Shenzhen, China \\
$^{2}$Shenzhen Malong Artificial Intelligence Research Center, Shenzhen, China\\
$^{3}$The University of Hong Kong\\
{\tt\small {\{terrencege,sheng,whuang,mscott\}@malong.com}}
}

\maketitle
\ificcvfinal\thispagestyle{empty}\fi

\begin{abstract}
	
Weakly-supervised instance segmentation aims to detect and segment object instances precisely, given image-level labels only.
Unlike previous methods which are composed of multiple offline stages, we propose Sequential Label Propagation and Enhancement Networks (referred as Label-PEnet) that progressively transform image-level labels to pixel-wise labels in a coarse-to-fine manner.
We design four cascaded modules including multi-label classification, object detection, instance refinement and instance segmentation, which are implemented sequentially by sharing the same backbone. The cascaded pipeline is trained alternatively with a curriculum learning strategy that generalizes labels from high-level images to low-level pixels gradually with increasing accuracy.
In addition, we design a proposal calibration module to explore the ability of classification networks to find key pixels that identify object parts, which serves as a post validation strategy running in the inverse order.
%
%
We evaluate the efficiency of our Label-PEnet in mining instance masks on standard benchmarks: PASCAL VOC 2007 and 2012. Experimental results show that Label-PEnet outperforms the state-of-the-art algorithms by a clear margin, and obtains comparable performance even with the fully-supervised approaches.
	
\end{abstract}

\section{Introduction}
Deep convolutional neural networks (CNNs) have made a series of breakthroughs in computer vision, by using large-scale manually-labeled data for training. By designing strong network architectures, CNNs can detect object locations and segment object instances precisely. However, the performance on object detection or segmentation will drop considerably due to lack of strong annotation provided at the object level or pixel level ~\cite{oquab2015object,diba2016weakly,ge2018multi,zhou2018weakly}, \ie when there are only image-level labels available.


To investigate the ability of CNNs to estimate pixel-wise labels when only image-level supervision is given, various weakly-supervised approaches have been developed for object detection or instance segmentation. A number of methods~\cite{bilen2016weakly,tang2017multiple,tang2018weakly} exploit a bottom-up approach to group pixels into proposals, and then evaluate the proposals repetitively in an effort to search exact object locations. Several algorithms dissect the classification process of CNNs in a top-down ~\cite{zhang2016top,lapuschkin2016analyzing} or bottom-up manner~\cite{zhou2016learning}, with the goal of generating seeds for instance segmentation~\cite{zhou2018weakly}. There are also some hybrid approaches that combine both bottom-up and top-down cues ~\cite{roy2017combining,ge2018multi}.


Existing weakly-supervised methods can achieve competitive results, but the performance is still significantly lower than that of fully-supervised counterparts. Although we can roughly identify an object using a classification network, it is particularly challenging to precisely infer pixel-wise labels from a classification model, even using multiple post-processing methods. This inspired us to re-think the ability of CNNs for various vision tasks, such as image classification, object detection and instance segmentation. We observed that full supervision with accurate annotations is the key to success. Therefore, the central issue for weakly-supervised detection and segmentation is to transfer image-level supervision to pixel-wise labels gradually and smoothly, in a coarse-to-fine manner by designing multiple cascaded modules.

The 2-D structure of convolutional kernels allows CNNs to grasp local information accurately, and enlarge the size of receptive fields gradually with the increase of convolutional layers, which enable the CNN model to memorize and classify objects accurately. Our goal is to enable CNNs to segment objects by just providing image-level labels. We design CNNs with such ability by introducing four new modules: (1) multi-label classification module, (2) object detection module, (3) instance refinement module, and (4) instance segmentation module, which are cascaded sequentially. \\

\noindent\textbf{Multi-Label Classification Module.} In this module, an image is first partitioned into a number of patches, generating a set of object proposals. We employ an unsupervised method, selective search~\cite{uijlings2013selective} or edge box~\cite{zitnick2014edge}, where pixels are organized by low-level statistics for generating object candidates.
%
Then a classification branch and a weight branch are incorporated to perform multi-label classification. In addition, we propose a proposal calibration module able to identify more accurate object locations and predict pixel-wise labels in object proposals.\\

\noindent\textbf{Object Detection Module.}
%
The rough object locations generated are used to train a standard object detection with Faster-RCNN~\cite{renNIPS15fasterrcnn}. But it can be unstable with direct training as we implemented.
%
Thus we explore object scores generated from the classification module to guide the training of current object detection, and infer object locations with the model during sequential learning. Similarly, we perform proposal calibration to identify pixels belonging to the corresponding objects, which further improve the detection accuracy. \\



\noindent\textbf{Instance Refinement Module.} With the generated object locations and instance masks, we perform instance segmentation using a standard Mask-RCNN~\cite{he2017mask}. However, current supervised information is still not accurate enough, so that we need to further explore object scores generated from the detection module to guide the training of current instance segmentation. Furthermore, a new instance branch is explored to perform instance segmentation, because the previous instance masks are generated based on individual samples, and can be rectified gradually with increasing accuracy when used as supervision.\\


\noindent\textbf{Instance Segmentation Module.} In this module, we obtain relatively strong supervision from the previous modules, which are used to guide the training of current instance segmentation, where final results are generated.\\

The main contributions of this work are summarized as:

First, we introduce Sequential Label Propagation  and  Enhancement Networks (Label-PEnet) for weakly-supervised instance segmentation. Our framework is composed of four cascaded modules that mine, summarize and rectify the appearance of objects repetitively. A two-stage training scheme is developed to train Label-PEnet effectively. It is an important step forward in exploiting the ability of CNNs to recognize objects from image level to pixel level, and thus boost up the performance of weakly-supervised instance segmentation.

Second, we propose a proposal calibration module to uncover the classification process of CNNs, and then mine the pixel-wise labels from image-level and object-level supervision. In this module, both top-down and bottom-up methods are explored and combined to identify object pixels with increasing accuracy.

Third, to validate the effectiveness of the proposed Label-PEnet, we conduct experiments on standard benchmarks: PASCAL VOC 2007 and PASCAL VOC 2012. Experimental results show that Label-PEnet outperforms state-of-art approaches by a clear margin, and obtains comparable performance even compared with fully supervised methods.

\begin{figure*}[ht]
	\centering
	\includegraphics[width=1.0\linewidth]{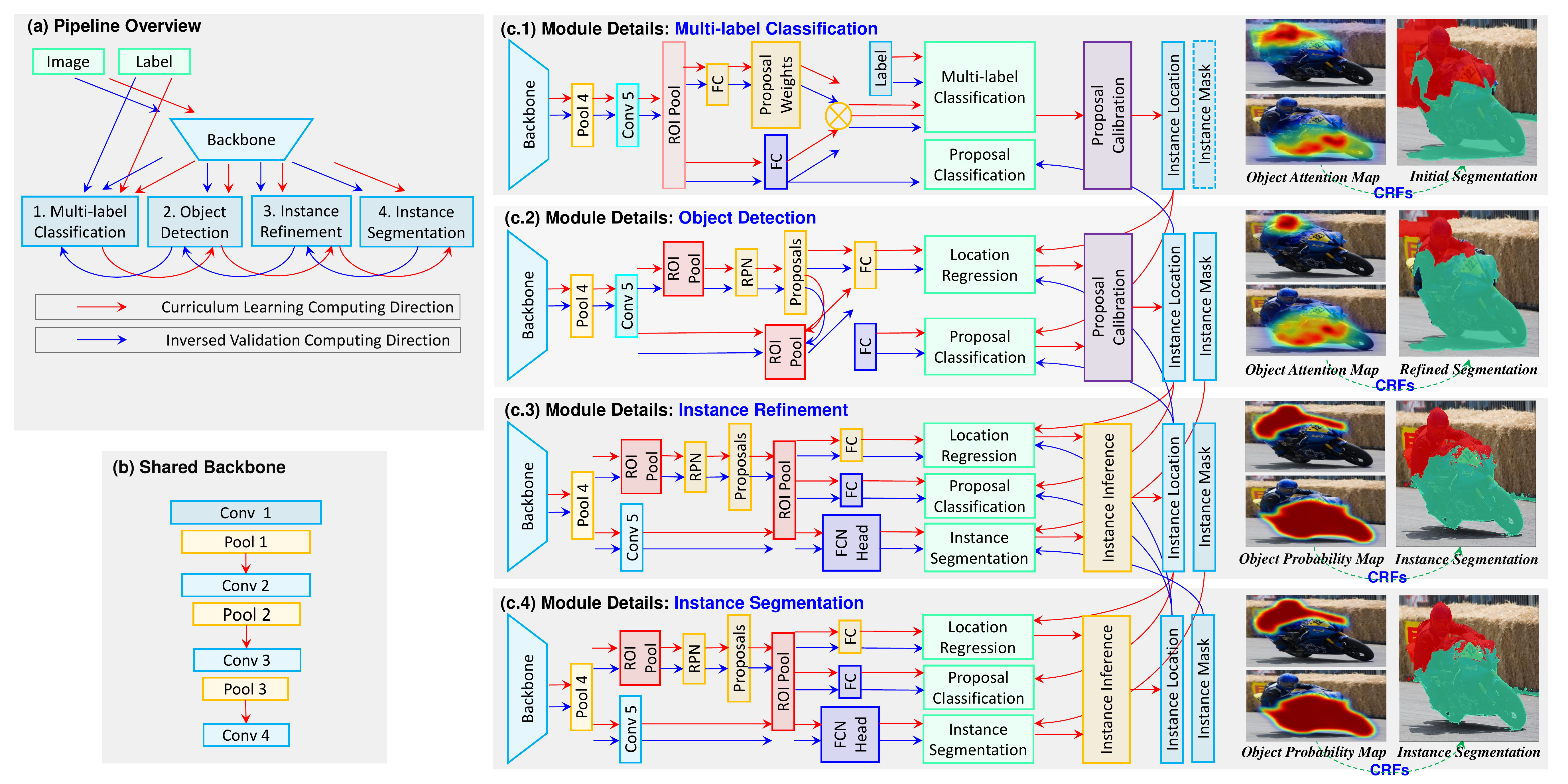}
\vspace{-6mm}
	\caption{The proposed Label-PEnet for weakly-supervised instance segmentation. (a) Overview: the training pipeline contains two different stages. One is curriculum learning stage which learns from image-level labels to pixel-wise labels. The other one learns in an inverse order to validate the results generated from the previous modules. (b) Shared backbone: the backbone is shared by all modules. (c) The details of different modules for multi-label classification, object detection, instance refinement, and instance segmentation. We develop a two-stage training scheme for learning Label-PEnet: a cascaded pre-training stage and a forward-backward learning stage. The backbone is fixed during the cascaded pre-training, and then is trained in forward-backward learning stage.}
	\label{Fig:WSIS}
\end{figure*}

\section{Related Work}
We briefly review the related studies on weakly-supervised object detection and segmentation, along with recent neural attention methods and applications of curriculum learning. \\

\noindent\textbf{Weakly-Supervised Object Detection and Segmentation.} Weakly-supervised object detection and segmentation is very challenging but is important to image understanding. They aim to locate and segment objects using image-level labels only~\cite{oquab2015object,diba2016weakly}. There are usually three kinds of methods: bottom-up manner, top-down manner, or the combination of two. For example, methods in \cite{oquab2015object,durand2016weldon,durand2017wildcat} treat the weakly-supervised object localization as a multi-label classification problem, and locate objects by using specific pooling layers. On the other hand, approaches in \cite{bilen2016weakly,tang2017multiple} extract and select object instances from images using selective search~\cite{uijlings2013selective} or edge boxes~\cite{zitnick2014edge}, and handle the weakly-supervised detection problem with multi-instance learning~\cite{dietterich1997solving}. The method in \cite{zhou2018weakly} attempted to find peaks in the class activation map, and then propagate the peaks to identify the corresponding object proposals generated by MCG~\cite{pont2017multiscale}. In this paper, we decompose the instance segmentation task into multiple simpler problems, and utilize the ability of CNNs to identify object pixels progressively. \\
\vspace{-0mm}

\noindent\textbf{Neural Attention.} Neural attention aims to understand the classification process of CNNs, and learn the relationship between the pixels in the input image and the neural activations in convolutional layers. Recent effort has been made to explain how neural networks work ~\cite{zhang2016top,bau2017network,lapuschkin2016analyzing}. In \cite{lapuschkin2016analyzing}, Lapuschkin \emph{et al.} extended a layer-wise relevance propagation (LRP)~\cite{bach2015pixel} to visualize inherent structured reasoning of deep neural networks. To identify the important regions producing final classification results, Zhang \emph{et al.} \cite{zhang2016top} proposed a positive neural attention back-propagation scheme, called excitation back-propagation (Excitation BP). Other related methods include Grad-CAM~\cite{selvaraju2017grad} and network dissection~\cite{bau2017network}.
Neural attention obtains pixel-wise class probabilities using image-level labels in a top-down manner on a well trained network. In our pipeline, we propose a forward network that computes pixel-wise class probability map for each individual proposal. This allows us to transfer image-level labels to pixel-wise ones, providing richer supervision for subsequent object detection and instance segmentation. \\

\noindent\textbf{Curriculum Learning.} Curriculum learning \cite{bengio2009curriculum} is set of machine learning methods that decompose a complicated learning task into multiple sub-tasks with gradually increasing learning difficulty. In \cite{bengio2009curriculum}, Yoshua \emph{et al.} described the concept of curriculum learning, and used a toy classification problem to show the advantage of decomposing a complex problem into multiple simpler ones. Various machine learning algorithms~\cite{sun2015robust,graves2017automated} follow a similar divide-and-conquer strategy in curriculum learning. Recently, Sheng \emph{et al.} \cite{Sheng2018} proposed CurriculumNet for large-scale weakly-supervised image classification.  CurriculumNet is able to learn high-performance CNNs from an image dataset containing a large amount of noisy images and labels, which were collected rawly from the Internet without any human annotation \cite{li2017webvision}.
\vspace{-0mm}
In this paper, we adopt this strategy to decompose the instance segmentation problem into multi-label image classification, object detection and instance segmentation sequentially. All the learning tasks in these modules are relatively simple by using the training data with the refined labels generated from previous stages

\begin{figure*}[ht]
	\centering

\subfigure[Object proposals]{\includegraphics[height=3.2cm,width=2.6cm]{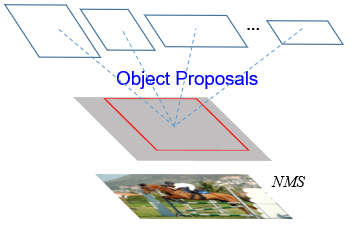}}
\subfigure[Excitation BP]{\includegraphics[height=3.2cm,width=2.6cm]{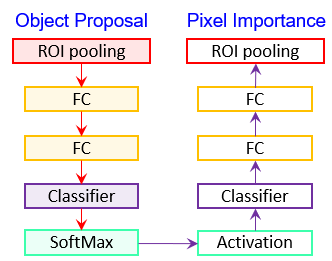}}
\subfigure[Proposal attention maps: ``Person"-``Horse"]{\includegraphics[height=3.2cm,width=5.5cm]{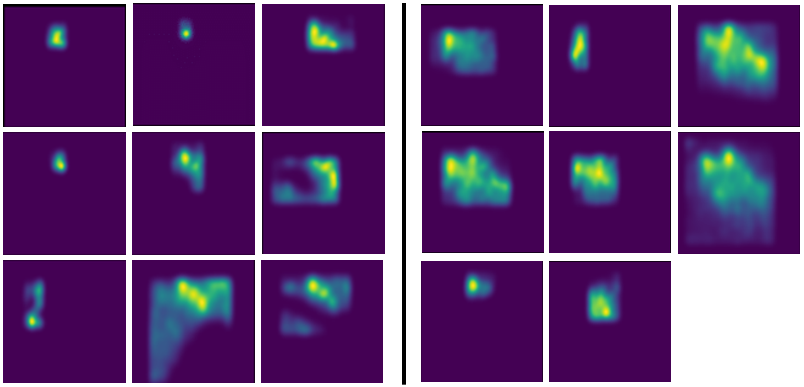}}
\subfigure[Object mask generation]{\includegraphics[height=3.2cm,width=6.5cm]{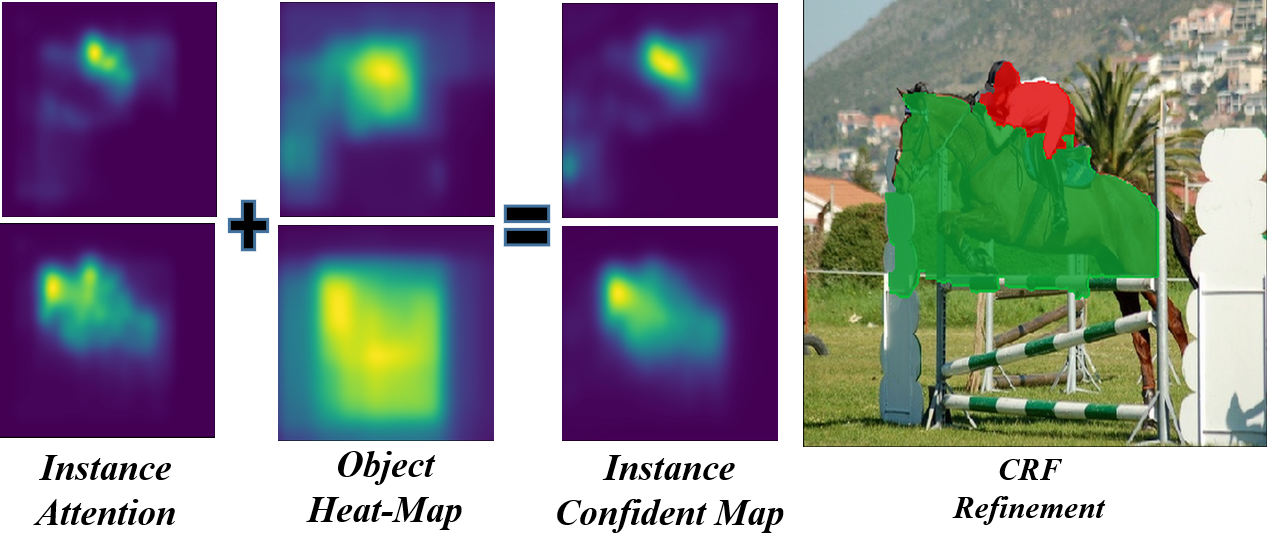}}
	\caption{Proposal calibration module. (a) Object proposals: all candidate object proposals suppressed by NMS are taken to generate a set of proposal attention maps. (b) Excitation BP: the process of Excitation BP implemented on each proposal. (c) The generated \emph{proposal attention maps} for ``Person" and ``Horse", and all proposal attention maps for a same instance are combined to generate a single \emph{instance attention map}. (d) Instance mask generation: instance attention map and object heat-map are combined to compute an instance confident map, where CRF~\cite{krahenbuhl2011efficient} is implemented to generate a final instance mask.}
\label{Fig:Proposal Dissection}
\end{figure*}

\section{Label-PEnet: Sequential Label Propagation and Enhancement Networks}

\subsection{Preliminary and Overview}
Given an image $\boldsymbol{I}$ associated with an image-level label $\boldsymbol{y}_I=[y^1, y^2, ..., y^\mathcal{C}]^T$, our goal is to estimate pixel-wise labels $\boldsymbol{Y}_I = [\boldsymbol{y}_1, \boldsymbol{y}_2, ..., \boldsymbol{y}_P]^T$ for each object instance. $\mathcal{C}$ is the number of object classes, $P$ is the number of pixels in $\boldsymbol{I}$. $y^l$ is a binary value, where $y^l = 1$ means the image $\boldsymbol{I}$ contains the $l$-th object category, and otherwise, $y^l = 0$. The label of a pixel $p$ is denoted by a $\mathcal{C}$-dimensional binary vector $\boldsymbol{y}_p$.
In this work, we propose a weakly-supervised learning approach for instance segmentation, which is inspired by the divide-and-conquer idea in curriculum learning \cite{bengio2009curriculum}. This allows us to train our model with increasingly stronger supervision which is learned automatically by propagating object information from image level to pixel level \emph{via} four cascaded modules: multi-label classification module, object detection module, instance refinement module, and instance segmentation module. The proposed Label-PEnet is described in Fig.~\ref{Fig:WSIS}.


\subsection{Multiple Cascaded Modules}\label{sec:detection}

\paragraph{Multi-Label Classification Module.} This module aims to generate a set of rough object proposals with corresponding class confident values and proposal weights, by just using image-level category labels.
To identify rough regions of objects, we exploit selective search~\cite{uijlings2013selective} to generate a set of object proposals $\boldsymbol{R} = (R_1, R_2, ..., R_n)$. These object candidates are then used as input to our multi-label classification module for collecting the proposals with higher confidence, and learning to identify the pixels which paly the key role in the classification task.

For an image $\boldsymbol{I}$ of $W \times H$, given a deep neural network $\phi_{d}(\cdot, \cdot ; \theta)$ with a convolutional stride of $\lambda_s$, we have convolutional feature maps with a spatial size of  $H/\lambda_s \times W/\lambda_s$ in the last convolutional layer.
Then ROI pooling~\cite{girshick2015fast} is performed on the convolutional feature maps to compute the features for each object proposals in $\boldsymbol{R}$,
resulting in $\left | \boldsymbol{R} \right |$ regional features for image $\boldsymbol{I}$. Two fully-connected layers are applied separately to the computed regional features, generating  classification results, ${\boldsymbol{x}^{c,1} \in {\mathbb{R}}^{\left | \boldsymbol{R} \right | \times  \mathcal{C}}}$, and weight vectors, ${\boldsymbol{x}^{p,1} \in {\mathbb{R}}^{\left | \boldsymbol{R} \right | \times  \mathcal{C}}}$, for the $\left | \boldsymbol{R} \right |$ object proposals. The proposal weights indicate the contribution of each proposal to the $\mathcal{C}$ categories in image-level multi-label classification. A softmax function is applied to normalize the weights as,

\begin{equation}
	\begin{aligned}
		\boldsymbol{w}^{p,1} _{ij} = \frac{e^{\boldsymbol{x}^{p,1}_{ij}}}{\sum_{i=1}^{\left | \boldsymbol{R} \right |} e^{\boldsymbol{x}^{p,1}_{ij}} }.
	\end{aligned}
	\label{eq:propsal weight normalization}
\end{equation}
where $\boldsymbol{x}^{p,1} _{ij}$ stands for the weight of the $i$-th proposal on the $j$-th class. We can have a normalized weight matrix ${\boldsymbol{w}^{p,1} \in {\mathbb{R}}^{\left | \boldsymbol{R} \right | \times  \mathcal{C}}}$. Then the final score for each proposal on different classes is  calculated by taking an element-wise product, $\boldsymbol{x}^1 = \boldsymbol{x}^{c,1} \odot \boldsymbol{w}^{p,1}$, and the final image-level multi-label classification results are computed by summing over all the proposals associated to each class, $s_c^1 = \sum_{i=1}^{\left | \boldsymbol{R} \right | } \boldsymbol{x}^1_{ic}$. This results in a final score vector for the input image $\boldsymbol{I}$, $\boldsymbol{s}^1 = \left [ s_1^1, s_2^1, ..., s_{\mathcal{C}}^1 \right ]$, indicating a confident value for each class. A probability vector $\boldsymbol{\hat{p}}^1 = \left [ \hat{p}_1^1, \hat{p}_2^1, ..., \hat{p}_{\mathcal{C}}^1 \right ]$ can be computed by applying a softmax function to $\boldsymbol{s}^1$, and the loss function for image-level multi-label classification is,
\begin{equation}
	\begin{aligned}
		\mathcal{L}_1(\boldsymbol{I},\boldsymbol{y}_I) = - \sum_{k=1}^{\mathcal{C}}  y^k \log \hat{p}^1_k.
	\end{aligned}
	\label{eq:loss 1}
\end{equation}

\paragraph{Proposal Calibration.}
The generated object proposals, with their classification scores, $\boldsymbol{x}^{c,1}$, are further processed by proposal calibration, which is a proposal refinement sub-module able to refine the generated proposals. The goal is to improve the prediction accuracy on object bounding boxes, and generate object masks, providing stronger and more accurate supervision for next modules.

Recent work of \cite{zhang2016top} introduces a new Excitation Back-Propagation (Excitation BP) able to generate a discriminative object-based attention map by using the predicted image-level class labels, which inspired us to compute an attention map for each proposal by using the predicted classification score. We explore a same network architecture as the classification module.
%
%
Specifically, given a proposal $R_i$, we apply a softmax function on its class prediction ${\boldsymbol{x}^{c,1}_i \in {\mathbb{R}}^{\mathcal{C}}}$ to have a normalized vector, $\boldsymbol{w}_i^{c,1}$, and predict an object class $c_i$ by using the highest value. Then we get a class activation vector,  ${\boldsymbol{a}^{c,1}_i \in {\mathbb{R}}^{\mathcal{C}}}$, by setting all other elements to 0, except for the $c_i$-th one in $\boldsymbol{w}_i^{c,1}$.
We perform the Excitation BP \cite{zhang2016top} in a feed forward manner from the classification layer to the ROI pooling layer by using the activation vector, generating a \emph{proposal attention map}, $\boldsymbol{A}_i$, for proposal $R_i$, as shown in Fig. \ref{Fig:Proposal Dissection}.
Then for the proposals with label $c$ in the image $\boldsymbol{I}$, we perform non-maximum suppression (NMS) by using the classification scores, $\boldsymbol{x}^{c,1}$, and generate an object candidate $R^c$ having the highest score. For those proposals (with label $c$) which are suppressed by $R^c$, we add their proposal attention maps by projecting them into the corresponding locations in the image, and generate a class-specific attention map $\boldsymbol{A}^c$, referred as \emph{instance attention map} for object class $c$, as shown in Fig. \ref{Fig:Proposal Dissection}.
Finally, we can compute a set of object instance attention maps: $\boldsymbol{A} = [\boldsymbol{A}^1, \boldsymbol{A}^2, ..., \boldsymbol{A}^\mathcal{C}] \in {\mathbb{R}}^ {\mathcal{C} \times H \times W}$, with a background map, $\boldsymbol{A}_0 = max(0, 1 - \Sigma_{l=1}^\mathcal{C}{{y^l}}\boldsymbol{A}_l)$.

We further compute an object heat-map for each instance class. The object heat-map for instance class $c$ is generated by computing pixel-wise sum over all proposals with class $c$, using the corresponding classification scores in $\boldsymbol{x}^{c,1}$. Then we combine instance attention maps and object heat-maps to generate final  \emph{instance confident maps}, where a conditional random field (CRF)~\cite{krahenbuhl2011efficient} is further implemented to segment object instances more accurately. This results in a set of segmentation masks, $\boldsymbol{S}^1 \in {\mathbb{R}}^ {\mathcal{K} \times H \times W}$, with corresponding object bounding boxes, $\boldsymbol{B}^1 \in {\mathbb{R}}^ {\mathcal{K} \times 4}$. Meanwhile, for each pair of bounding box and segmentation mask, we simply use the classification score of the identified object candidate (e.g., $R^c$) as a weight, obtaining the predicted instance weights $\boldsymbol{W}^1 \in {\mathbb{R}}^ \mathcal{K}$ which are used to guide the training of next object detection module.
\vspace{-3mm}

\paragraph{Object Detection Module.} With the generated proposal bounding boxes $\boldsymbol{B}^1$ and the corresponding weights $\boldsymbol{W}^1$, we train a standard object detection model by using them as ground truth.  The main difference is that we provide a learned weight for each generated proposal during training. By following Faster-RCNN~\cite{ren2015faster}, we sample positive and negative proposals around a ground truth bounding box, and each proposal sampled has a same weight with the corresponding ground truth.
%
Then the optimization objective of region proposal network (RPN) is modified as,
\begin{equation}
	\begin{aligned}
		L\left ( w_i,  t_i \right )_{rpn} = & \frac{1}{N_{rpn}} \sum _{i} L_{obj}(w_i,  w_i^*) \\
		& + \lambda  \frac{1}{N_{rpn}} \sum _{i} w_i^* L_{reg}(t_i,  t_i^*),
	\end{aligned}
	\label{eq:loss 1}
\end{equation}
where $N_{rpn}$ is the number of candidate proposals, $w_i$ is the predicted object score, $t_i$ is the predicted location offset, $w_i^*$ is the proposal weight, $t_i^*$ is the pseudo object location, $\lambda$ is a constant value. $L_{obj}$, $L_{cls}$  and $L_{reg}$ are the binary object or non-object loss, classification loss, and bounding-box regression loss respectively. For the RCNN part, the optimization objective is computed as,
\begin{equation}
	\begin{aligned}
		L\left ( p_i,  t_i \right )_{rcnn} = & \frac{1}{N_{rcnn}} \sum _{i} w_i^* L_{cls}(p_i,  p_i^*) \\
		& + \lambda  \frac{1}{N_{rcnn}} \sum _{i} w_i^* L_{reg}(t_i,  t_i^*).
	\end{aligned}
	\label{eq:loss 1}
\end{equation}
where $p_i$ is the classification score, and $p_i^*$ indicates the object class. $N_{rcnn}$ is the number of proposals generated by RPN, and $L_{cls}$ is the classification loss.
On the head of Faster-RCNN architecture, we perform proposal calibration to refine object proposals, which is similar to that of  multi-label classification module. This enables the model to generate dense proposal attention maps. In inference, multiple object candidates can be generated for multiple labels, which are different from the proposal calibration in classification module that outputs one candidate for each label. Finally, we can obtain multiple instance marks, $\boldsymbol{S}^2$, with corresponding bounding boxes, $\boldsymbol{T}^2$, and weights, $\boldsymbol{W}^2 \in {\mathbb{R}}^ \mathcal{J}$, where $\mathcal{J}$ is the number of object instances.
%

\vspace{-3mm}

\paragraph{Instance Refinement Module.} With the generated instance masks $\boldsymbol{S}^2$ and object bounding boxes $\boldsymbol{T}^2$, we can train an instance segmentation task having a joint detection branch and mask branch similar to that of Mask R-CNN ~\cite{he2017mask}. In this module, we implement instance inference for dense pixel-wise prediction rather than proposal calibration, by following the feed forward inference as~\cite{he2017mask}.
Object instances are learnt and modeled in the module by collecting part of the information hidden in the results generated from previous modules. We perform object instance segmentation with the learned weights $\boldsymbol{W}^2$, and our training process follows that of Mask-RCNN ~\cite{he2017mask}.
%
%
As in the proposal calibration, object masks affiliated with the predicted object location are summed together to generate a new instance confident map. Similarly, we perform CRF~\cite{krahenbuhl2011efficient} to obtain more accurate results of instance segmentation.
\vspace{-3mm}

\paragraph{Instance Segmentation Module.} In this module, image-level labels have been successfully transferred into dense pixel-wise labels. We perform standard instance segmentation in a fully supervised manner, by simply following the training strategies implemented in the instance refinement module. Final results can be generated during inference.

\begin{figure*}[ht]
	\centering
	\includegraphics[height=4.5cm,width=16.5cm]{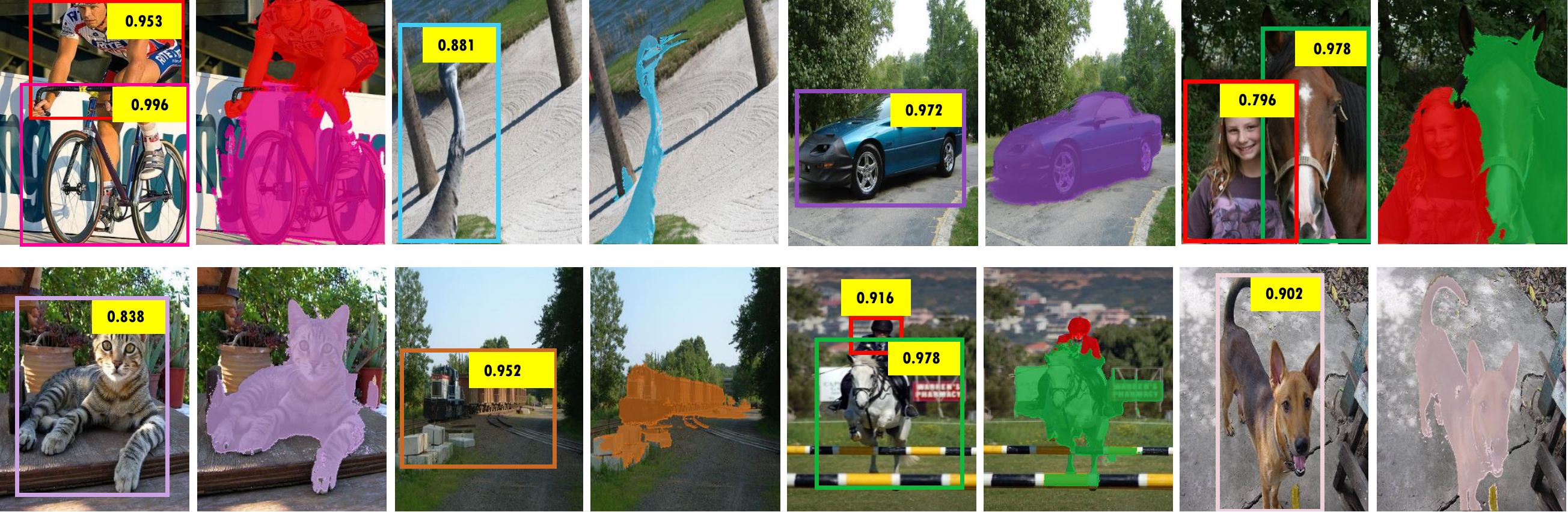}
\vspace{-1mm}
	\caption{Instance detection and segmentation results on Pascal VOC 2012 (the first row) and Pascal VOC 2007 (the second row). The proposals with the highest confidence are selected and visualized. The segmentation results are post-processed by CRF~\cite{krahenbuhl2011efficient}.}
	\label{Fig:instance segmentation}
\end{figure*}

\subsection{Training with Label Propagation}
\begin{table*}[t]\small
	\setlength{\abovecaptionskip}{10pt}
	\setlength{\belowcaptionskip}{-10pt}
	\begin{center}
		\resizebox{1\textwidth}{!}
		{
			\begin{tabular}{@{}lcccccccccccccccccccccc@{}}
				\toprule
				method                                & aero         &bike  &bird          &boat          &bottle        &bus           &car  &cat           &chair &cow          &table &dog          &horse &mbike         &person &plant         &sheep         &sofa         &train &tv             &mAP   \\ \midrule
				OM+MIL+FRCNN\cite{li2016weakly}       & 54.5         &47.4  &41.3          &20.8          &17.7          &51.9          &63.5 &46.1          &21.8  &57.1         &22.1  &34.4         &50.5  &61.8          &16.2   &29.9          &40.7          &15.9         &55.3  &40.2           &39.5  \\
				HCP+DSD+OSSH3\cite{jie2017deep}       & 54.2         &52.0  &35.2          &25.9          &15.0          &59.6          &67.9 &\textbf{58.7} &10.1  &\textbf{67.4}&27.3  &37.8         &54.8  &\textbf{67.3} &5.1    &19.7          &52.6          &43.5         &56.9  &62.5           &43.7  \\
				OICR-Ens+FRCNN\cite{tang2017multiple} &\textbf{65.5} &67.2  &47.2          &21.6          &22.1          &68.0  &\textbf{68.5}&35.9          &5.7   &63.1 &\textbf{49.5} &30.3         &64.7  &66.1          &13.0   &25.6          &50.0          &57.1         &60.2  &59.0           &47.0  \\
				MEFF+FRCNN\cite{ge2018multi}          &64.3          &68.0  &\textbf{56.2} &\textbf{36.4} &23.1          &68.5          &67.2 &64.9          &7.1   &54.1         &47.0  &\textbf{57.0}&69.3  &65.4  &\textbf{20.8}  &23.2          &50.7          &59.6 &\textbf{65.2} &57.0           &51.2  \\ \midrule
				Multi-label Cls Module$^{\dagger}$    & 41.2         &42.0  &6.5           &17.1          &7.1           &54.1          &40.5 &8.5           &17.3  &33.0         &13.2  &10.3         &24.4  &54.0          &5.5    &7.5           &20.0          &39.2         &49.9  &47.3           &26.9  \\
				Object Det Module$^{\dagger}$         & 49.1         &61.3  &24.8          &15.9          &46.9          &58.9          &25.3 &17.7          &23.3  &41.8         &28.9  &42.4         &67.1  &25.3          &6.7    &50.4          &40.9          &62.4         &50.4  &42.3           &39.1  \\
				Instance Ref Module$^{\dagger}$       & 62.3         &68.3  &47.2          &27.9          &53.8          &69.1          &39.9 &41.9          &25.9  &56.5         &40.1  &53.0         &70.0  &44.9          &13.3   &53.5          &51.1          &68.6         &60.9  &45.2           &49.7  \\
				Instance Seg Module$^{\dagger}$       & 63.8         &69.0  &47.9          &35.3  &\textbf{56.1}         &68.9          &41.5 &42.7          &25.9  &58.3         &44.3  &52.5         &70.3  &44.4          &13.8   &\textbf{56.9} &52.9          &70.0         &62.3  &49.9           &51.3 \\ \midrule
				Multi-label Cls Module$^{\ddagger}$   & 42.4         &43.8  &8.9           &18.7          &6.5           &55.7          &42.0 &10.0          &18.3  &34.3         &14.5  &11.4         &24.8  &56.2          &3.7    &9.1           &22.1          &40.5         &51.1  &46.5           &28.0  \\
				Object Det Module$^{\ddagger}$        & 51.2         &63.0  &28.8          &17.5          &51.1          &60.3          &28.9 &20.7          &25.9  &41.0         &31.2  &46.4         &68.1  &27.1          &6.0    &50.9          &43.6          &65.8         &50.6  &40.3           &40.3  \\
				Instance Ref Module$^{\ddagger}$      & 63.2         &67.5  &48.3          &29.8          &54.8          &70.4          &40.9 &42.6  &\textbf{27.9} &55.0         &41.5  &54.3         &70.0  &43.2          &15.3   &55.4          &52.4          &69.0         &62.2  &46.8           &50.5  \\
				Instance Seg Module$^{\ddagger}$      & 65.7 &\textbf{69.4} &50.6          &35.8          &55.5  &\textbf{71.9}         &43.6 &45.3          &27.5  &58.5         &45.4  &55.4 &\textbf{71.7} &45.8          &18.2   &56.6  &\textbf{56.1} &\textbf{72.0}        &64.6  &51.4   &\textbf{53.1} \\
				\bottomrule
			\end{tabular}
		}
	\end{center}
	\vspace{-3mm}
	\caption{Average precision (in \%) of weakly-supervised methods on PASCAL VOC 2007 $detection~test$ set. $^{\dagger}$ stands for the results of the cascaded pre-training. $^{\ddagger}$ stands for the results of the recurrent mixed fine-tuning.}
	\label{voc07 detection}
\end{table*}

\begin{table*}[t]\small
	\setlength{\abovecaptionskip}{10pt}
	\setlength{\belowcaptionskip}{-10pt}
	\begin{center}
		\resizebox{1\textwidth}{!}
		{
			\begin{tabular}{@{}lcccccccccccccccccccccc@{}}
				\toprule
				method                                & aero                & bike          &bird        & boat         & bottle       & bus          & car           &cat          & chair        & cow          & table        & dog          & horse        & mbike        & person        & plant        & sheep        & sofa         & train        & tv           & mAP  \\ \midrule
				OICR-VGG16\cite{tang2017multiple}     & 67.7                & 61.2          &41.5        & 25.6         & 22.2         & 54.6         & 49.7          &25.4         & 19.9         & 47.0         & 18.1         & 26.0         & 38.9         & 67.7         & 2.0           & 22.6         & 41.1         & 34.3         & 37.9         & 55.3         & 37.9 \\
				WSDDN+context\cite{diba2016weakly}    & 64.0                & 54.9          &36.4        & 8.1          & 12.6         & 53.1         & 40.5          &28.4         & 6.6          & 35.3         & 34.4         & 49.1         & 42.6         & 62.4  &\textbf{19.8}         & 15.2         & 27.0         & 33.1         & 33.0         & 50.0         & 35.3 \\
				HCP+DSD+OSSH3+NR\cite{jie2017deep}    & 60.8                & 54.2          &34.1        & 14.9         & 13.1         & 54.3         & 53.4          &58.6         & 3.7          & 53.1         & 8.3          & 43.4         & 49.8         & 69.2         & 4.1           & 17.5         & 43.8         & 25.6         & 55.0         & 50.1         & 38.3 \\
				OICR-Ens+FRCNN\cite{tang2017multiple} &\textbf{71.4} &\textbf{69.4}         &55.1        & 29.8         & 28.1         & 55.0         & 57.9          &24.4         & 17.2  &\textbf{59.1}        & 21.8         & 26.6         & 57.8         & 71.3         & 1.0           & 23.1  &\textbf{52.7}        & 37.5         & 33.5  &\textbf{56.6}        & 42.5 \\
				MEFF+FRCNN\cite{ge2018multi}          & 71.0                & 66.9  &\textbf{55.9}&\textbf{33.8}        & 24.0         & 57.6  &\textbf{58.0} &\textbf{61.4}        & 22.5         & 58.4         & 19.2  &\textbf{58.7}        & 61.9  &\textbf{75.0}        & 11.2          & 23.9         & 50.3         & 44.9         & 41.3         & 54.3         & 47.5 \\ \midrule
				Multi-label Cls Module$^{\ddagger}$   & 37.1                & 40.0          &5.9         & 11.7         & 5.5          & 48.3         & 40.5          &7.0          & 16.3         & 29.2         & 9.9          & 8.3          & 19.3         & 51.1         & 3.0           & 6.1          & 17.0         & 36.3         & 46.4         & 39.1         & 23.9  \\
				Object Det Module$^{\ddagger}$        & 49.2                & 57.0          &25.1        & 13.9         & 49.5         & 53.3         & 25.3          &15.9         & 20.0         & 36.5         & 29.1         & 42.1         & 60.9         & 22.9         & 5.5           & 43.5         & 37.8         & 63.4         & 48.7         & 35.8         & 36.8  \\
				Instance Ref Module$^{\ddagger}$      & 57.9                & 65.5          &43.9        & 26.9         & 50.9         & 64.7         & 35.9          &38.7         & 22.8         & 50.9         & 38.9         & 50.9         & 65.5         & 39.5         & 13.6          & 52.9         & 48.9         & 65.7         & 57.9         & 41.9         & 46.7  \\
				Instance Seg Module$^{\ddagger}$      & 60.8                & 65.4          &46.2        & 31.4  &\textbf{50.3} &\textbf{68.3}        & 40.7          &39.9  &\textbf{25.3}        & 52.8  &\textbf{43.4}        & 53.9  &\textbf{68.2}        & 40.8         & 15.9   &\textbf{53.1}        & 50.0  &\textbf{68.1} &\textbf{59.8}        & 49.0  &\textbf{49.2} \\
				\bottomrule
			\end{tabular}
		}
	\end{center}
	\vspace{-3mm}
	\caption{Average precision (in \%) of weakly-supervised methods on PASCAL VOC 2012 $detection~test$ set.}
	\label{voc12 detection}
\end{table*}

\begin{table*}[t]\small
	\setlength{\abovecaptionskip}{10pt}
	\setlength{\belowcaptionskip}{-10pt}
	\begin{center}
		\resizebox{1\textwidth}{!}
		{
			\begin{tabular}{@{}lcccccccccccccccccccccc@{}}
				\toprule
				method                                & aero         & bike  & bird         & boat         & bottle        & bus          & car         & cat         & chair        & cow          & table        & dog         & horse        & mbike        & person        & plant        & sheep        & sofa         & train        & tv           & mCorLoc  \\ \midrule
				OICR-VGG16\cite{tang2017multiple}     & 81.7         & 80.4  & 48.7         & 49.5         & 32.8          & 81.7         & 85.4        & 40.1 &\textbf{40.6}        & 79.5         & 35.7         & 33.7        & 60.5         & 88.8         & 21.8          & 57.9  &\textbf{76.3}        & 59.9         & 75.3  &\textbf{81.4}        & 60.6     \\
				WSDDN-Ens\cite{diba2016weakly}        & 68.9         & 68.7  & 65.2         & 42.5         & 40.6          & 72.6         & 75.2        & 53.7        & 29.7         & 68.1         & 33.5         & 45.6        & 65.9         & 86.1         & 27.5          & 44.9         & 76.0         & 62.4         & 66.3         & 66.8         & 58.0     \\
				OM+MIL+FRCNN\cite{li2016weakly}       & 78.2         & 67.1  & 61.8         & 38.1         & 36.1          & 61.8         & 78.8        & 55.2        & 28.5         & 68.8         & 18.5         & 49.2        & 64.1         & 73.5         & 21.4          & 47.4         & 64.6         & 22.3         & 60.9         & 52.3         & 52.4     \\
				HCP+DSD+OSSH3\cite{jie2017deep}       & 72.2         & 55.3  & 53.0         & 27.8         & 35.2          & 68.6         & 81.9        & 60.7        & 11.6         & 71.6         & 29.7         & 54.3        & 64.3         & 88.2         & 22.2          & 53.7         & 72.2         & 52.6         & 68.9         & 74.4         & 54.9     \\
				OICR-Ens+FRCNN\cite{tang2017multiple} & 85.8  &\textbf{82.7} & 62.8         & 45.2  &\textbf{43.5}  &\textbf{84.8} &\textbf{87.0}       & 46.8        & 15.7         & 82.2  &\textbf{51.0}        & 45.6        & 83.7  &\textbf{91.2}        & 22.2   &\textbf{59.7}        & 75.3         & 65.1         & 76.8         & 78.1         & 64.3     \\
				MEFF+FRCNN\cite{ge2018multi}          & 88.3         & 77.6  & 74.8         & 63.3         & 37.8          & 78.2         & 83.6        & 72.7        & 19.4         & 79.5         & 46.4  &\textbf{78.1}       & 84.7         & 90.4         & 28.6          & 43.6         & 76.3         & 68.3  &\textbf{77.9}        & 70.6         & 67.0     \\ \midrule
				Label-PEnet                              &\textbf{89.8} & 82.6  &\textbf{75.3} &\textbf{65.7} & 39.2          & 80.2         & 81.6 &\textbf{77.7}       & 18.4  &\textbf{82.7}        & 49.3         & 75.0 &\textbf{86.9}        & 85.9  &\textbf{30.7}         & 49.6         & 75.3  &\textbf{71.5}        & 76.1         & 70.6  &\textbf{68.2}    \\
				\bottomrule
			\end{tabular}
		}
	\end{center}
	\vspace{-3mm}
	\caption{CorLoc (in \%) of weakly-supervised methods on PASCAL VOC 2007 $detection~trainval$ set.}
	\label{voc07 localization}
\end{table*}

\begin{table*}[t]\small
	\setlength{\abovecaptionskip}{10pt}
	\setlength{\belowcaptionskip}{-10pt}
	\begin{center}
		\resizebox{1\textwidth}{!}
		{
			\begin{tabular}{@{}lcccccccccccccccccccccc@{}}
				\toprule
				method                                        & aero        & bike        & bird        & boat & bottle        & bus         & car  & cat         & chair & cow         & table & dog         & horse        & mbike & person        & plant & sheep        & sofa        & train & tv           & mCorLoc  \\ \midrule
				OICR-VGG16\cite{tang2017multiple}             & 86.2        & 84.2        & 68.7        & 55.4 & 46.5          & 82.8        & 74.9 & 32.2        & 46.7  & 82.8        & 42.9  & 41.0        & 68.1         & 89.6  & 9.2           & 53.9  & 81.0         & 52.9        & 59.5  &\textbf{83.2} & 62.1     \\
				WSDDN+context\cite{diba2016weakly}            & 78.3        & 70.8        & 52.5        & 34.7 & 36.6          & 80.0        & 58.7 & 38.6        & 27.7  & 71.2        & 32.3  & 48.7        & 76.2         & 77.4  & 16.0          & 48.4  & 69.9         & 47.5        & 66.9  & 62.9         & 54.8     \\
				HCP+DSD+OSSH3+NR\cite{jie2017deep}            & 82.4        & 68.1        & 54.5        & 38.9 & 35.9          & 84.7        & 73.1 & 64.8        & 17.1  & 78.3        & 22.5  & 57.0        & 70.8         & 86.6  & 18.7          & 49.7  & 80.7         & 45.3 &\textbf{70.1} & 77.3         & 58.8     \\
				OICR-Ens+FRCNN\cite{tang2017multiple}  &\textbf{89.3}&\textbf{86.3}       & 75.2        & 57.9 &\textbf{53.5}  & 84.0 &\textbf{79.5}& 35.2 &\textbf{47.2} &\textbf{87.4}& 43.4  & 43.8        & 77.0         & 91.0  & 10.4   &\textbf{60.7} &\textbf{86.8} & 55.7        & 62.0  & 84.7         & 65.6     \\
				MEFF+FRCNN\cite{ge2018multi}                  & 88.0        & 81.6        & 75.8        & 60.9 & 46.2          & 85.3        & 75.3 & 76.5        & 47.2  & 85.4        & 47.7  & 74.3        & 87.8  &\textbf{91.4} & 21.6          & 55.3  & 77.9  &\textbf{68.8}       & 64.9  & 75.0         & 69.4     \\ \midrule
				Label-PEnet                                      & 89.1        & 84.3 &\textbf{78.8}&\textbf{63.2}& 47.9   &\textbf{88.7}       & 76.8 &\textbf{77.2}& 46.3  & 87.2 &\textbf{50.4} &\textbf{78.9}&\textbf{91.8} & 90.1  &\textbf{25.7}  & 56.3  & 78.5         & 66.3        & 69.9  & 78.3  &\textbf{71.3}    \\
				\bottomrule
			\end{tabular}
		}
	\end{center}
	\vspace{-3mm}
	\caption{CorLoc (in \%) of weakly-supervised methods on PASCAL VOC 2012 $detection~trainval$ set.}
	\label{voc12 localization}
\end{table*}

\begin{table*}[t]\small
	\setlength{\abovecaptionskip}{10pt}
	\setlength{\belowcaptionskip}{-10pt}
	\begin{center}
		\resizebox{1\textwidth}{!}
		{
			\begin{tabular}{@{}lccccccccccccccccccccccc@{}}
				\toprule
				method                               &bg            &aero          &bike          &bird          &boat          &bottle        &bus           &car           &cat           &chair         &cow           &table         &dog           &horse         &mbike          &person        &plant         &sheep         &sofa          &train         &tv            &mIoU          \\ \midrule
				SEC\cite{kolesnikov2016seed}         &83.5          &56.4          &28.5          &64.1          &23.6          &46.5          &70.6          &58.5          &71.3          &\textbf{23.2} &54.0          &28.0          &68.1          &62.1          &70.0           &55.0          &38.4          &58.0          &39.9          &38.4          &48.3          &51.7          \\
				FCL\cite{roy2017combining}           &85.7          &58.8          &30.5          &67.6          &24.7          &44.7          &74.8          &61.8          &73.7          &22.9          &57.4          &27.5          &71.3          &64.8          &\textbf{72.4}  &57.3          &37.0          &60.4          &42.8          &42.2          &\textbf{50.6} &53.7          \\
				TP-BM\cite{kim2017two}               &83.4          &62.2          &26.4          &\textbf{71.8} &18.2          &49.5          &66.5          &63.8          &73.4          &19.0          &56.6          &35.7          &69.3          &61.3          &71.7           &\textbf{69.2} &39.1          &66.3          &\textbf{44.8} &35.9          &45.5          &53.8          \\
				AE-PSL\cite{wei2017object}           &-             &-             &-             &-             &-             &-             &-             &-             &-             &-             &-             &-             &-             &-             &-              &-             &-             &-             &-             &-             &-             &55.7          \\
				MEFF\cite{ge2018multi}               &86.6          &72.0          &30.6          &68.0          &\textbf{44.8} &46.2          &73.4          &56.6          &73.0          &18.9          &63.3          &32.0          &70.1          &72.2          &68.2           &56.1          &34.5          &\textbf{67.5} &29.6          &60.2          &43.6          &55.6          \\
				MCOF-VGG16\cite{wang2018weakly}      &85.8          &\textbf{74.1} &23.6          &66.4          &36.6          &\textbf{62.0} &\textbf{75.5} &\textbf{68.5} &\textbf{78.2} &18.8          &\textbf{64.6} &29.6          &\textbf{72.5} &61.6          &63.1           &55.5          &37.7          &65.8          &32.4          &\textbf{68.4} &39.9          &56.2          \\ \midrule
			    Label-PEnet                             &\textbf{88.3} &73.5          &\textbf{33.2} &70.1          &43.2          &49.1          &74.7          &57.0          &75.9          &20.7          &64.3          &32.3          &72.3          &\textbf{74.1} &71.1           &57.4          &38.3          &69.1          &31.2          &61.1          &45.2          &\textbf{57.2} \\
				\bottomrule
			\end{tabular}
		}
	\end{center}
	\vspace{-3mm}
	\caption{Comparisons of weakly-supervised semantic segmentation methods on PASCAL VOC 2012 $segmentation~test$ set.}
	\label{voc12 test sgementation}
\end{table*}

To better train multiple sequential models and avoid local minima, we initialize the backbone network with an ImageNet pre-trained model. The training is implemented sequentially by using the output of previous module, with gradually enhanced supervision.
We develop a two-stage training process containing cascaded pre-training and forward-backward learning with curriculum.
\vspace{-3mm}

\paragraph{Cascaded Pre-Training.} The backbone networks are fixed during cascaded pre-training. We pre-train four cascaded modules sequentially, from multi-label classification to instance segmentation. When the training of current module is converged, with model outputs well regularized and refined, such outputs are then used as supervision for the next module.
With the cascaded pre-training, we decompose a weakly-supervised instance segmentation task into four sequential sub-tasks where image-level supervision is propagated gradually and efficiently to dense pixel-wise predictions. 
\vspace{-3mm}

\paragraph{Forward-Backward Learning with Curriculum} Training four sequential models is challenging, because networks might get into local minima easily with sequential label propagation. To overcome this problem, we propose a forward-backward learning method by leveraging curriculum learning, which has a forward curriculum learning phase and backward validation phase, as shown in Fig.~\ref{Fig:WSIS}.
In the forward curriculum learning, the four modules are trained sequentially where the supervised information is enhanced gradually.  While in the backward validation, training is performed in an inverse order. The backward validation starts from instance segmentation module, where we just perform inference at the module, and generate object locations and instance masks for instance refinement module. Then the instance refinement module is trained in a fully supervised manner, providing object locations for object detection module.
In multi-label classification module, we set the proposals, which have an overlap of $>\beta$ (= 0.5) with the objects detected by the detection module, with a label of the corresponding objects or background. Then we perform single-label classification on these proposals, and at the same time, keep training multi-label classification task.

\section{Experimental Results}
Our methods were implemented using Caffe~\cite{jia2014caffe} and run on a NVIDIA TITAN RTX GPU with 24GB memory. The parameters of object detection and instance segmentation modules are the same with Faster R-CNN~\cite{renNIPS15fasterrcnn} and Mask R-CNN~\cite{he2017mask}. Several examples are illustrated in Fig. \ref{Fig:instance segmentation}.

\subsection{Network Structures}
\paragraph{Backbone Network.} The backbone network is based on VGG-16, where the layers after $relu4\_3$ are removed. As shown in Fig~\ref{Fig:WSIS}, only the first four convolutional blocks are preserved. All the parameters are initialized from an ImageNet pre-trained model.
\vspace{-4mm}

\paragraph{Multi-label Classification Module.} Following the backbone network, the fifth convolution block contains $conv5\_1$, $conv5\_2$, and $conv5\_3$. We set dilations in the three layers to 2. The feature stride $\lambda_s$ at layer $relu5\_3$ is 8. A ROI pooling~\cite{girshick2015fast} is added to generate a set of $512 \times 7 \times 7$ feature maps, followed by $fc6$ and $fc7$ layers.
The classification branch and proposal weight branch are initialized randomly using a Gaussian initializer as in~\cite{he2016deep}.
\vspace{-4mm}

\paragraph{Object Detection Module.} As in multi-label classification module, dilations in $conv5\_1$, $conv5\_2$, and $conv5\_3$ are set to 2. The RPN~\cite{renNIPS15fasterrcnn} contains three convolutional layers which are all initialized with Gaussian distributions with 0 mean and standard deviation of 0.01. It generates proposals where ROI pooling~\cite{girshick2015fast} is conducted on the feature maps $relu5\_3$. A proposal classification branch and a bounding box regression branch are presented by following two fully-connected layers $fc6$ and $fc7$.
\vspace{-4mm}

\paragraph{Instance Refinement Module and Instance Segmentation Module.} The two modules have the same network architecture, which contains an object detection part and an instance segmentation part. The object detection part is similar to that of object detection module, with only one difference that the RPN and ROI pooling are computed on the feature maps of the $pool4$ layer, not the $relu5\_3$. For the instance segmentation part, we adopt the atrous spatial pyramid pooling as that of DeepLab V3 \cite{chen2017rethinking} after layer $relu5\_3$, with dilations set to  $[1, 2, 4, 6]$.

\subsection{Implementation Details}
\paragraph{Cascaded Pre-Training.} In the cascaded pre-training stage, we train the four cascaded modules in a forward order, but keep the parameters in the backbone network fixed. For data augmentation, we use five image scales, \{480, 576, 688, 864, 1024\} (for the shorter side), and horizontal flip, and cap the longer side at 1,200. The mini-batch size for SGD is set to 2, and the learning rate is set to 0.001 in the first 40K iterations, and then is decreased to 0.0001 in the following 10K iterations. The weight decay is 0.0005, with a momenta of 0.9. These settings are used in all the four modules. We start training the next module only when the training of previous one is finished. Selective Search (SS)~\cite{uijlings2013selective} is adopted in the multi-label classification module to generate about 1,600 object proposals per image. For RPN in object detection module and instance segmentation module, we follow~\cite{renNIPS15fasterrcnn} to use 3 scales and 3 aspect ratios, yielding $k = 9$ anchors at each sliding position. The sizes of convolutional feature maps after ROI pooling in the detection branch and segmentation branch are $7 \times 7$ and $14 \times 14$.
\paragraph{Forward-Backward Learning with Curriculum.} As shown in Fig~\ref{Fig:WSIS}, there are two sub-stages for training: a forward curriculum learning stage and an inverse  backward validation stage, which are implemented alternatively at each iteration. All layers with learnable parameters are trained in an end-to-end manner. The training starts from the cascaded pre-trained model, with a learning rate of 0.0001 in the following 80K iterations. 
In inference, an image with original size is used as input.
\vspace{-0mm}

\subsection{Weakly Supervised Object Detection}
\noindent\textbf{Dataset and Evaluation.} We evaluate the performance of weakly-supervised object detection on Pascal VOC 2007 and Pascal VOC 2012~\cite{everingham2015pascal}. Both datasets were divided into train, val and test sets. The trainval sets (with 5011 images for 2007 and 11540 images for 2012) are used for training, where only image-level labels are used, without any bounding box information or pixel-wise annotation. By following the standard Pascal VOC protocol, the performance on object detection is evaluated on the test sets by using mAP, while object localization accuracy~\cite{deselaers2012weakly} is measured on the trainval sets by using  CorLoc (the correct localization).

\vspace{-0mm}

\noindent\textbf{Results.} 
Object detection results on Pascal VOC 2007 and 2012 are reported in Table \ref{voc07 detection} and \ref{voc12 detection}. Object localization results are presented in Table \ref{voc07 localization} and \ref{voc12 localization}. On Pascal VOC 2007 test set, our method achieves the highest mAP (53.1\%), with at least 1.9\% higher than recent methods, including MEFF~\cite{ge2018multi}, OICR\cite{tang2017multiple} and HCP+DSD+OSSH3\cite{jie2017deep}. It also has the highest mAP (49.2\%) among all weakly-supervised algorithms on Pascal VOC 2012 test set,
with 1.7\% higher than the latest results reported in~\cite{ge2018multi}. For object localization, our performance are highly competitive among the state-of-the-art results, by achieving  68.2\% and 71.3\% on Pascal VOC 2007 and 2012, respectively, which are 1.2\% and 1.9\% higher than the previous best results.

\subsection{Weakly-Supervised Semantic Segmentation}
\noindent\textbf{Dataset and Evaluation.} Pascal VOC 2012 dataset~\cite{everingham2015pascal} is the standard benchmark for the task of weakly-supervised semantic segmentation. It contains 21 classes with 10,582 images for training (including VOC 2012 training set and additional data annotated in \cite{hariharan2011semantic}), 1,449 images for validation and 1,456 for test. Only image-level labels are used for training. We do not use any additional data annotated in \cite{hariharan2011semantic}, and report the results on the test set in Table \ref{voc12 test sgementation}.

\noindent\textbf{Results.} 
As shown in Table~\ref{voc12 test sgementation}, our method achieves a mean IoU of 57.2\%, and outperforms the previous state-of-the-art AE-SPL\cite{wei2017object} and MCOF~\cite{wang2018weakly} by 1.6\% and 1\% respectively. Compared with recent algorithms, including AE-SPL\cite{wei2017object}, F-B~\cite{saleh2016built}, FCL~\cite{roy2017combining}, and SEC~\cite{kolesnikov2016seed}, our Label-PEnet cast the semantic segmentation problem into multiple easier tasks, which allows us to propagate high-level image labels to pixel-wise labels gradually with enhanced accuracy.

\begin{table}[!t]\small
	\setlength{\abovecaptionskip}{10pt}
	\setlength{\belowcaptionskip}{-10pt}
	\begin{center}
		\resizebox{0.4\textwidth}{!}
		{
			\begin{tabular}{@{}lccccccccccccccccccccccc@{}}
				\toprule
				method                              & mAP$^r_{0.25}$ & mAP$^r_{0.5}$  & mAP$^r_{0.75}$  & ABO            \\ \midrule
				PRM-VGG16    \cite{zhou2018weakly}  & -              & 22.0           & -               & -              \\
				PRM-ResNet50 \cite{zhou2018weakly}  & 44.3           & 26.8           & 9.0             & 37.6           \\ \midrule
				Label-PEnet                           & \textbf{49.1}  & \textbf{30.2}  & \textbf{12.9}   & \textbf{41.4}  \\
				\bottomrule
			\end{tabular}
		}
	\end{center}
\vspace{-3mm}
	\caption{Comparisons of weakly-supervised instance segmentation methods on Pascal VOC 2012 $validation$ set.}
	\label{voc12_instance}
\end{table}

\subsection{Weakly-Supervised Instance Segmentation}
\noindent\textbf{Dataset and Evaluation.} We follow the experimental settings in~\cite{zhou2018weakly} by using Pascal VOC 2012 dataset~\cite{everingham2015pascal} for weakly-supervised instance segmentation. Experimental results are evaluated with mAP$^r$ at IoU threshold of 0.25, 0.5 and 0.75, and the Average Best Overlap (ABO)~\cite{pont2015boosting}. We report results on the test set in Table \ref{voc12_instance}.

\noindent\textbf{Results.} We use VGG16 as our backbone, and report the performance in the term of four metrics, while most existing methods used ResNet50. Only PRM-VGG16 applied VGG16 and obtained a mAP$^r_{0.5}$ of 22.0\%. Obviously, our method outperforms PRM-VGG16 by 8.2\% on  mAP$^r_{0.5}$. Even compared with PRM-ResNet50, our method can obtain large improvements on all four metrics.


\subsection{Evaluation on Individual Modules}
We further compare the effect of each individual modules on the test set of Pascal VOC 2007 detection, as shown in Table~\ref{voc12 detection}. In the cascaded pre-training, the multi-label classification can only have a mAP of 26.9\%, which is improved to 39.1\% when we refine object locations with the proposal calibration module and detection module. Furthermore, instance refinement module further improves the object detection results considerably by 10.6\%, reaching to 49.7\%. Finally, the instance segmentation module can achieve a mAP of 51.3\%.
%
%
The results suggest that with more accurate results provided as guidance and supervision, the object detection results can be improved gradually and significantly with four cascaded modules. When we perform the forward-backward learning, our Label-PEnet can have a mAP of 53.1\%, which is 1.8\% higher than that of the cascaded pre-training, and also outperforms previous methods, such as MEFF+FRCNN \cite{ge2018multi} and OICR-Ens+FRCNN \cite{tang2017multiple}.

\section{Conclusions}
We have presented new Sequential Label Propagation and Enhancement Networks (referred as
Label-PEnet) for weakly-supervised object detection and instance segmentation. Label-PEnet is able to progressively transform image-level labels to pixel-wise predictions in a coarse-to-fine manner, by designing four cascaded modules, from multi-label classification, object detection, instance refinement to instance segmentation.
%
In addition, we design a proposal calibration module to explore the ability
of classification CNNs to identify key pixels of objects, which further improves detection and segmentation accuracy. Our Label-PEnet is evaluated on the standard benchmarks for weakly-supervised object detection and segmentation, where it outperformed the state-of-the-art methods by a clear margin.


{\small
\bibliographystyle{ieee_fullname}
\bibliography{egbib}
}

\end{document}